\documentclass{article}

\usepackage{PRIMEarxiv}

\usepackage{times}
\usepackage{epsfig}
\usepackage{graphicx}
\usepackage{amsmath}
\usepackage{hyperref}
\usepackage{amssymb}

\usepackage{subfig}  
\usepackage{algorithm}  
\usepackage{algpseudocode} 
\usepackage{breqn} 
\usepackage{multirow} 

\newcommand{\prob}{\mathbb{P}}

\title{Discrete Control in Real-World Driving Environments using Deep Reinforcement Learning
}

\author{
  Avinash Amballa\\
 Bosch \\
  \texttt{amballaavinash@gmail.com} \\
   \And
  Advaith P \\
 Indian Institute of Technology Hyderabad \\
\texttt{ep19btech11004@iith.ac.in} \\
  \And
  Pradip Sasmal \\
 Indian Institute of Technology Jodhpur \\
  \texttt{psasmal@iitj.ac.in} \\
\And
  Sumohana Channappayya \\
 Indian Institute of Technology Hyderabad \\
  \texttt{sumohana@ee.iith.ac.in} \\
}

\begin{document}
\maketitle

\begin{abstract}
Training self-driving cars is often challenging since they require a vast amount of labeled data in multiple real-world contexts, which is computationally and memory intensive. Researchers often resort to driving simulators to train the agent and transfer the knowledge to a real-world setting. Since simulators lack realistic behavior, these methods are quite inefficient. To address this issue, we introduce a framework (perception, planning, and control) in a real-world driving environment that transfers the real-world environments into gaming environments by setting up a reliable Markov Decision Process (MDP). We propose variations of existing Reinforcement Learning (RL) algorithms in a multi-agent setting to learn and execute the discrete control in real-world environments. Experiments show that the multi-agent setting outperforms the single-agent setting in all the scenarios. We also propose reliable initialization, data augmentation, and training techniques that enable the agents to learn and generalize to navigate in a real-world environment with minimal input video data, and with minimal training. Additionally, to show the efficacy of our proposed algorithm, we deploy our method in the virtual driving environment TORCS.
\end{abstract}


\section{Introduction}
\label{sec:1}

Self-driving vehicles \cite{DBLP:t1} \cite{NVIDIA}  have increased dramatically over the last few years due to the advances in Machine learning (ML) and Reinforcement Learning (RL). Self-driving vehicles will have an enormous economic impact over the coming decades. Creating models that meet or exceed a human driver's ability could save thousands of lives a year. One of the current industry-standard methods to train driving agents is to train supervised learning algorithms with a huge amount of labeled data \cite{NVIDIA}. The major disadvantage of this method is that it depends on the volume and quality of the training data. Different kinds of network architectures were developed to increase the capacity of network models, and increasingly larger datasets are being gathered today \cite{kim2018textual} \cite{ettinger2021large} \cite{49052}. The collection and annotation of large-scale datasets are time-consuming and expensive. The other widely used method is to train a reinforcement learning algorithm in a simulator \cite{Torcs} \cite{Carla}. Even though modern simulators are well designed, they fail to represent the challenges in the real-world scenario since the simulator environment is very different from real-world scenarios, which lack realistic behavior, graphics, and rendering these simulators computationally expensive. So training agents on these simulators and using them in the real world can be an issue. 


In contrast of training the agent on the simulator, or training using supervised data (large collection of video data), we train the agent on real-world un-labeled videos (from the vision sensor as input) using proposed RL algorithms. This method overcomes the challenges discussed earlier by taking the best from the two methods discussed (unlabeled real-world videos +  RL setting). However, comparing the RL agent trained on these unlabeled videos with the supervised setting is not a recommended one, since our aim is not to achieve state-of-the-art performance, instead, we propose a realistic and suitable framework for training RL agents. It is also easy to render this framework since it uses unlabeled video frames which can be easily collected through a vision sensor. Our proposed method can be widely used in scenarios where we only have access to the real video input(unlabeled), such as in environments with limited or no sensors. The only input we need is the video source that a human driver records. However, performance will improve with more information, such as sensor values. \\

We make the following contributions to this paper: 
\begin{enumerate}
    \item Introduce a framework that transforms the real-world driving scenario into a game scenario using perception (process and transform the environment), planning (predict the actions), and control (execute the actions) by adapting a reliable MDP. 
    \item
    Propose reliable data augmentation, initialization, and training methods  so that the agent learns and generalizes to navigate in a real-world  environment with minimal input video data, and with minimal training. 
    \item Modify existing RL algorithms from a single-agent to adapt to the multi-agent setting. This setting helps in faster learning.
    \item Adapt the proposed method for the virtual simulated environment TORCS with appropriate modifications in the pre-processing steps. 
\end{enumerate}

   \begin{figure*}[t]
    \centering
      \includegraphics[width=\linewidth]{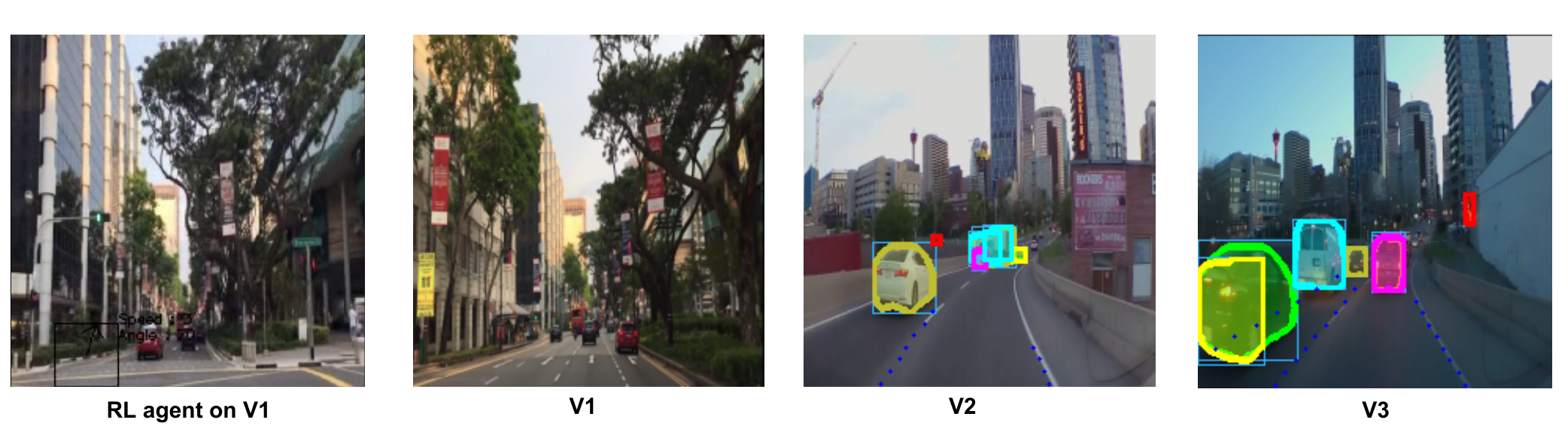}
    \caption{Real-world driving videos. Figures 1.1 and 1.2 represent the RL agent on a frame in the video V1, a frame in the video V1 respectively. Figures 1.3 and 1.4 represent lane markings, bounding boxes, and segmentation masks on a frame in the videos V2 and V3 respectively.}
    \label{fig:Figure 1}
    \end{figure*}

\section{Related Work} \label{sec:2}

\textbf{Training Driving Agents:} 
There are two approaches to training driving agents. The first one of which is using a supervised learning approach. This method directly maps the input image to driving action predictions. ALVIN \cite{ALVINN} is an early attempt to use a shallow neural network to map images input to driving action. Recently,  NVIDIA \cite{NVIDIA} proposed a similar work. However, it uses a CNN to map raw pixels directly from a single front-facing camera to steering commands. The system was only given images without lane markings, and the system was able to automatically learn end-to-end the internal representations of the necessary processing steps, such as detecting useful road features with only the human steering angle as the training signal.

The other widely used method is to train a Reinforcement Learning (RL) agent in a simulator. The RL framework consists of the environment in which the agent observes the current state $s$, executes an action based on a learned policy $\pi (a|s)$, and transits to a new state $s^\prime$. This transition from state $s$ to $s^\prime$ returns a reward $r$ to the agent. The agent's goal is to maximize the cumulative rewards by finding an optimal policy $\pi^*$. Many works \cite{DBLP:t1} \cite{t12} \cite{t13} have been done on the TORCS, an open-source racing simulator \cite{Torcs}. In \cite{DBLP:t1}, they used a DDPG algorithm for continuous control. A custom reward with appropriate sensor information from TORCS \cite{Torcs} was used to train this agent. Kamar et al. \cite{t12} uses deep Q-learning as an agent. A reward function promoting longitudinal velocity while penalizing transverse velocity and divergence from the track center was used to train the agent. In \cite{t13}, two reinforcement learning algorithms, DDPG and DQN, are compared and analyzed in the TORCS \cite{Torcs} simulation environment. The agent's goal was to finish the track by controlling the car. 

 Fayjie et al. \cite{U5} used  DQN as the agent, which was trained in a simulated urban environment. The agent was given two types of sensor data as input: a camera sensor and a laser sensor in front of the car. Chen et al. \cite{U7} focused on making a robust agent for complex urban environments. This paper provides a framework for model-free deep reinforcement learning in challenging urban autonomous driving scenarios. They also designed specific input representation and visual encoding to capture the low-dimensional latent states. These simulator environments are very different from real-world scenarios. The work \cite{sim_to_real} proposed a novel realistic translation network to make a model trained in a virtual environment workable in the real world.
 
 Apart from TORCS \cite{Torcs}, there are many other simulators. CARLA is one of them. It is an open-source simulator for autonomous driving research. CARLA supports flexible specifications of sensor suites and environmental conditions. Dosovitskiy et al. \cite{Carla} used CARLA to study the performance of three approaches to autonomous driving: a classic modular pipeline, an end-to-end model trained via imitation learning, and an end-to-end model trained via RL.

\section{Notations}
If a variable at $t^{th}$ timestep is denoted by $I$, we denote the variable at $(t+1)^{th}$ timestep with $I^\prime$ . Throughout this paper, we assume that the roads are uni-directional, which means that vehicles move only in one direction.  For ease of representation, we assume that the angle is always multiplied with a factor $\left(\frac{\pi}{180^\circ}\right)$ to convert into radians before passing into any trigonometric expressions.

\textbf{Coordinate system:} Throughout this paper, we represent the vehicles in 3D coordinate system (real-world system)  where $x,y,z$ represent coordinates in the direction of the vehicle's width, length, and height respectively. We often ignore the $z$ coordinate since the vehicle's motion is restricted to the $XY$ plane, however, we use the notion of $XZ$ plane to represent the plane of bounding boxes.

\section{Perception: From Video Environment to Game Environment}
     In this section, we introduce a perception block that converts real-world driving environments into gaming environments by setting up an MDP ($\mathcal{S, A, P, R}$) with state space $S$, action space $A$, state transition matrix as $\prob$, and reward function $R$. We represent the RL agent at the bottom of the video as a black box for visualization and inference Figure:\ref{fig:Figure 1}. An optimal policy would have this RL agent operate in the same manner as a human who recorded the video. 

    \subsection{Action Space}  \label{sec4:.1}
    Throughout this paper, we study the discrete action space setting. The discrete action space consists of two components, speed $f$ and angle $\theta$. We define frame rate(the rate at which frames within a video move) as speed  and the direction in which the agent moves is the angle. However, note that the term speed refers to the magnitude and does not have direction. The direction factor is taken care of by angle $\theta$. The action space  $f \in \{0,1,\dots f_{max}\}$, where $f_{max}$ is the maximum speed and action space $\theta \in \{10^\circ,20^\circ,30^\circ,\dots,170^\circ\}$. The brake condition is by default when the speed $f = 0$. Currently, our only controllable parameters are speed $f$ and steering angle $\theta$ of the agent so that it follows a lane and does not collide with other objects. Hence, the action space $\mathcal{A} \triangleq [f, \theta]$. Define $a_{t} \in \mathcal{A}, a_t = [f_{t}, \theta_{t}]$ be the action predicted in ${t}^{th}$ timestep. Once the action is taken, the previously predicted action becomes the current action i.e., $a_{t}$.

    
\subsection{State Space} \label{sec4:.2}

     A well-defined state space is a necessary and challenging task in the RL setting. However, we have access only to the video frames and do not have access to inputs such as sensor values, which are typically available in a simulator environment. In order to efficiently make use of input frames, we carry out necessary pre-processing steps on every frame. These steps include object detection, segmentation, lane detection, and depth map estimation.
     
    \begin{figure}[t]
    \centering
      \includegraphics[width=0.8\linewidth]{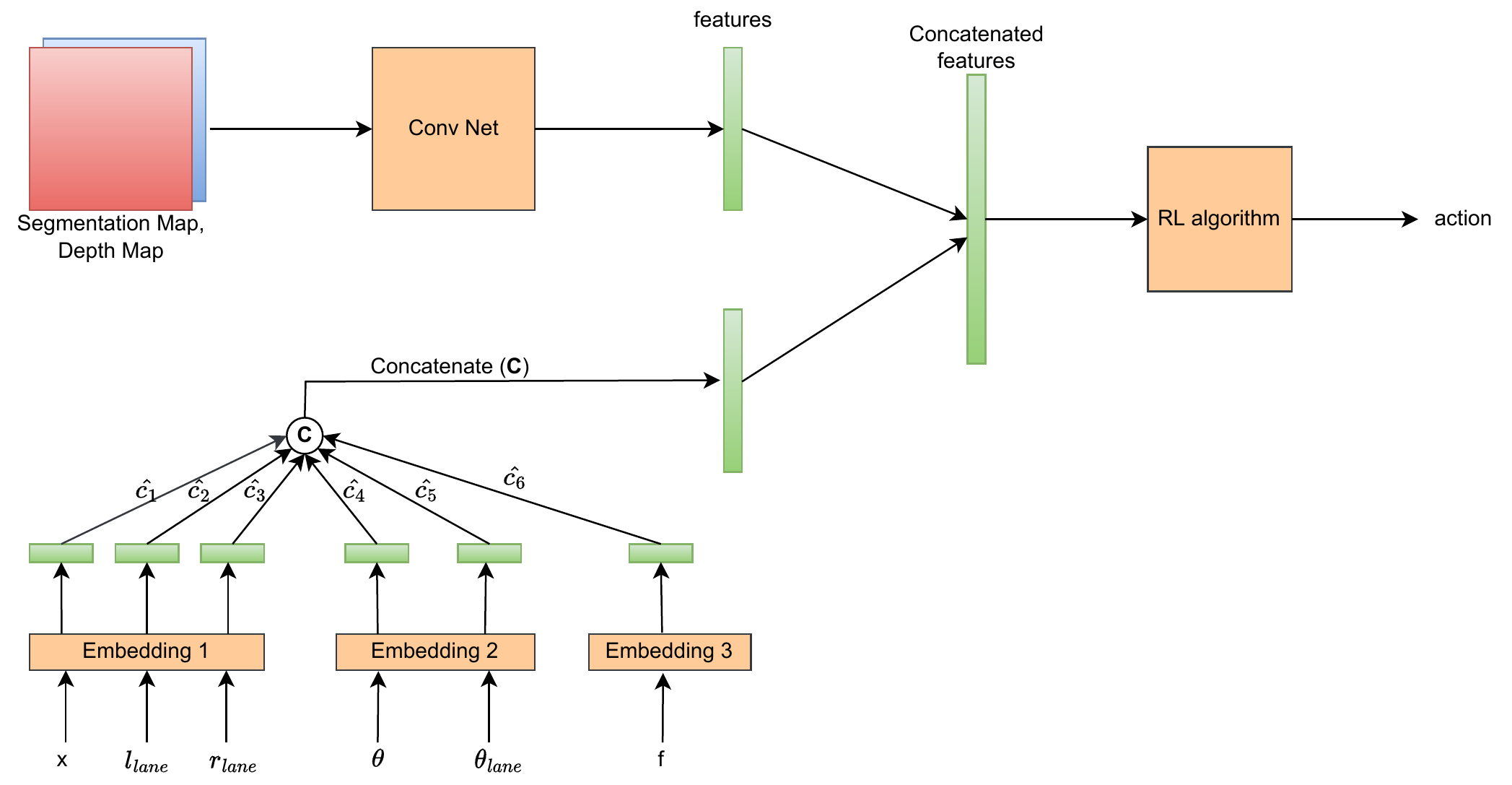}
    \caption{Representation of State Space $S_t$ at $t^{th}$ timestep}
    \label{fig:Figure 2}
    \end{figure}
    
    The object detection and segmentation model uses a Mask-RCNN architecture \cite{he2017mask}. Mask-RCNN is an improved version of the R-CNN where R-CNN generates region proposals based on selective search and then processes each proposed region one at a time using a CNN to output an object label and its bounding box. The lane detection model is based on \cite{qin2020ultra}, which was trained on the cleaned \cite{pan2018spatial}  dataset. This method treats the lane detection process as a row-based selecting problem using global features. Since we don't have access to stereo inputs from left and right cameras, we use the depth map model based on \cite{godard2019digging} where the model is a standard fully convolutional U-Net trained using self-supervised learning methods. This model estimates the depth map from a monocular image. 
     
     

     We pre-process every frame in the video and store the bounding box coordinates of all the objects within the current frame. After segmentation, we output the masks for certain objects relevant to the driving scenario. Instance segmentation is preferred over semantic segmentation because it helps the agent recognize each object as a distinct thing and respond accordingly. Similarly, for every frame, we store the lane coordinates for all the detected lanes and the depth map of the frame. Using the lane coordinates section:\ref{sec:4.3}   obtained, we calculate the lane angle $\theta_{lane}$, lane left extreme $l_{lane}$, lane right extreme $r_{lane}$.
    
    We define the state space Figure:\ref{fig:Figure 2} as vector $\mathcal{S} \triangleq $ [features from concatenated depth map and segmentation map obtained from the current frame, current agent's center position $(x,y)$, current speed $f$, current angle $\theta$, current lane left extreme $l_{lane}$, current lane right extreme $r_{lane}$, current lane angle $\theta_{lane}$ ]. However, in the agent position, we take only the $x$ coordinate of the agent since we make the $y$ coordinate constant. Concatenating depth maps and segmentation maps are based on the intuition that we get attention to the nearer and more important objects in the input image while ignoring the rest. This agrees with how we have defined our reward function, which focuses on the maximum intersection, which is essentially the objects with lesser depth. For feature extraction, we use a custom CNN $\footnote{We use CNN with 4 Conv Layers of 8, 16, 32, 64 channels with kernel size 5$\times$5, each followed by a 2$\times$2 max pool and fully connected layer at the end}$. 

      As the number of features from concatenated depth map and segmentation map in the state space $\mathcal{S}$ are higher when compared to other features, a simple concatenation might bias the outcome towards these features. However, to give importance to the other features as well, we introduce weighted concatenation Figure:\ref{fig:Figure 2} of the features where weights are determined using attention \cite{bahdanau2014neural}. We map the rest of the features in the state space to a higher dimension with the help of an embedding and apply attention to these embedding layers. \footnote{We use embedding size $e = 10$ with attention parameter $d = 10$. The parameters $V$, $W$, and $b$ are learned during training} 

     Let the embedding size be $e$ and the attention parameter be $d$. Let $ E \in$ $\mathbb{R}^{1\times e}$  be the embedding network, $V \in$ $\mathbb{R}^{d\times1}$,  $W \in$ $\mathbb{R}^{d\times e}$, $b \in$ $\mathbb{R}^{d\times1}$. We calculate
     \begin{equation} 
	\begin{split}
	c_i &= V^T \cdot \tanh(W \cdot E(S[i])^T+b) \forall i = \{1,2,\dots,6\}, \\
    \hat{c}_i & = \frac{\text{exp}(c_i)}{\sum_{j=1}^{6}\text{exp}(c_j)}
	\end{split}
	\label{eq:eq1}
	\end{equation}
    The final weighted concatenated state space at $t^{th}$ timestep $S_t \in \mathcal{S}$ is $S_t$ =  [features from concatenated depth map and segmentation map obtained from frame $I_t$, $\hat{c}_1 \cdot x_{t-1}$, $ \hat{c}_2\cdot f_{t-1}$, $\hat{c}_3\cdot \theta_{t-1}$,  $\hat{c}_4\cdot {l_{lane_{t}}}$, $\hat{c}_5\cdot {r_{lane_{t}}}$, $\hat{c}_6\cdot {\theta_{lane_{t}}}$  ]

\subsection{Reward Formulation } $\label{sec:4.3}$
Real-world video footage represents the actions performed by a human agent, which we consider optimal action since achieving human-level performance is the ultimate goal. Hence, the goal of the optimal agent is to mimic this human behavior by avoiding collisions with other objects and moving within a lane by maximizing reward. To address this, the reward formulation includes both object detection and lane detection rewards. Both these rewards work together to ensure that a positive reward is given to the agent in the case of optimal behavior and penalizes the agent in the other case. Let  $S_t = s$ be the state of the environment in $t^{th}$ timestep. Using the policy, agent predicts action $a_t = a^\prime$ i.e., $[f^\prime,  \theta^\prime]$ and executes the action and moves to state $S_{t+1} = s^\prime $. Define the intersection of the agent with an object as the common area between the bounding boxes of the agent and the object detected by Mask-RCNN. Define $M$ be the maximum of intersections between the agent and other objects detected by Mask-RCNN in $t^{th}$ timestep. Similarly, $M^\prime$ is the maximum of intersections between the agent and other objects detected by Mask-RCNN in $(t+1)^{th}$ timestep.\\
  
    We define three rules for reward formulation: 
    \begin{enumerate}
        \item  If the intersection of the agent with other objects is high, speed should be reduced, otherwise, speed can be high. 
        \item If the intersection of the agent with other objects is high, shift the lane, otherwise, stay in the same lane to encourage stability. 
        \item  Stay away from other objects by avoiding the intersection.
    \end{enumerate}
    
     Based on these three rules, we define our reward function for transition ($s,a^\prime,s^\prime$) as:  
    \begin{algorithmic}                 

    \If{$M < $ threshold }
        \State   reward $= \frac{\alpha \cdot (f^\prime-f)}{f_{max}}  + \beta \cdot reward_{lane} - \delta \cdot M^\prime$
              
    \Else :     
          \State  reward $= -\frac{\alpha\cdot(f^\prime-f)}{f_{max}} - \beta \cdot reward_{lane} - \delta \cdot M^\prime$
                   
    \EndIf
     
     \State where $reward_{lane} = $  $\mu\cdot |\cos(\theta^\prime-\theta^\prime_{lane}) | - \mu \cdot |\sin(\theta^\prime-\theta^\prime_{lane}) |  - \nu\cdot\frac{ |l^\prime-x^\prime |}{w}$, here $\theta^\prime_{lane}$ is the lane angle and $l^\prime$ is the center ($x$ coordinate) of any two closest lanes to agent in $(t+1)^{th}$ timestep, $w$ is frame size.
     
	\end{algorithmic}

    The agent makes angle $\theta$ with the $x-axis$ Figure:\ref{fig:Figure 4} i.e., the line perpendicular to the agent plane makes an angle of $\theta$ anticlockwise with $x-axis$. However, the object detection bounding boxes always make an angle of $90^\circ$ with $x-axis$. So to define the intersection area of the agent with an object in the frame, we project the agent plane parallel to the $XZ$-plane, i.e., the plane parallel to the rest of the objects. We then define the area of intersection between an agent with height, width $(H1, W1)$ and an object with height, and width $(H2, W2)$ as  $(H1 \cdot W1 \cdot \cos(\theta-90^\circ)  \cap H2 \cdot W2)$. Since finding this intersection is difficult, we approximate this  with $(H1 \cdot W1 \cap H2 \cdot W2 ) \cdot \cos(\theta-90^\circ) $. Given the coordinates, the intersection area of two parallel rectangles is straightforward. The same procedure applies in calculating $M^\prime$.
    
    \textbf{Lane coordinates, Lane angle, Lane extremes: }
    The lane detection algorithm outputs the lane coordinates for each and every frame. For each of the lanes in a frame, we calculate the angle of a lane curve using the start point $(p_1,q_1)$ and endpoint $(p_2,q_2)$ of a lane by approximating it with a straight line joining the start point and end point i.e., $\arctan (\frac{q_2-q_1}{p_2-p_1})$. We define $\theta_{lane}$ for a frame as the arithmetic mean of all lane curve angles in that frame. We take the lane extremes $l_{lane}$ and $r_{lane}$ as the left-most coordinate(left extreme) and right-most(right extreme) coordinate of the lanes detected within the frame respectively. In scenarios where there were no lanes  detected, we take  $\theta_{lane} = 90^\circ$ and lane extremes as $0.2w$, and $0.8w$ respectively. 
         
    \textbf{ Done Condition: }
    We terminate the episode when either the agent reaches the end of the video or the agent collides \footnote{we choose the threshold to be 0.2 and collision threshold to be 0.5} with any other object or agent moves out of the entire road or the agents stay at rest for more than some threshold time. Collision happens if max intersection $M >$ collision threshold. We define the agent movement out of the entire road if the agent is left to left-most lane $l_{lane}$ or right to right-most lane $r_{lane}$.

\subsection{Setting up Markov Decision Process (MDP)}  \label{sec:4.4}

In previous sections, we have defined our state space $\mathcal{S}$, action space $\mathcal{A}$ and Reward $\mathcal{R}$. 
We can write $S_t = g(I_t, x_{t-1}, a_{t-1})$ where $I_t$ is the current frame, $g$ is some arbitrary function since features from concatenated depth map and segmentation map of $I_t$,  ${l_{lane}}_t$, ${r_{lane}}_t$ and ${\theta_{lane}}_t$ depend on $I_t$ itself. Clearly $I_t$ depends on $I_{t-1}$  and $x_{t-1}$ depends on $x_{t-2}$ as shown in section:\ref{sec:6.1}, hence $S_t = g(S_{t-1}$,$a_{t-1}$)

 We took $a_t$ to be only dependent on current state $S_t$ i.e., $a_t$ = $\pi(S_t)$ for some policy $\pi$. This means the current state $S_t$ is alone sufficient to determine the action $a_t$.  The intuition behind this assumption is that in a real-world driving environment, the agent must make decisions based on the immediate current state, so providing the agent with knowledge about the previous states could lead the agent to learn a policy that might not be the optimal one. 
When a stochastic policy $\pi$ is combined with the Markov decision process, the action for each state is frozen and the system behaves like a Markov reward process. 

Under learned policy $\pi$, \\
    \begin{dmath} 
	\prob (S_t=s^\prime|S_{t-1}=s, a_{t-1}=a) = 
    \prob (S_t=s^\prime|S_{t-1}=s, a_{t-1}=\pi(s)) \nonumber
	 \label{eq:eq2}
	\end{dmath}

We model the game scenario as a first-order finite horizon MDP $\mathcal{(S, A, P, R)}$ with state space $S$, action space $A$, state transition matrix as $\prob_{ss^{1}}$, reward function $R$ in addition to a learned policy $\pi$. However, note that the way we have defined our state-space makes an implicit dependency of action $a_t$ on states $S_{t}, S_{t-1}, S_{t-2}, \ldots, S_{0}$


\section{Planning: Algorithm}  \label{sec:5}
The issue with the real-world driving environment is that frames captured one after another are highly correlated. This can be a significant problem for neural networks because performing numerous iterations of gradient descent on correlated and similar inputs may cause them to over-fit or enter a local minimum. Another key challenge is that learning an optimal policy requires a large number of samples in the driving environments. To overcome these challenges, we adopt a multi-agent setting and the below-mentioned training procedure. 

Since we deal with continuous state space, and discrete action space, we use  Vanilla Policy Gradient \cite{sutton1999policy}, Dueling DQN \cite{pmlr-v48-wangf16}, and Advantage Actor-Critic (A2C) algorithm \cite{mnih2016asynchronous} and adapt them to the multi-agent setting. The reason to choose these algorithms is that they cover all the existing families(settings) in RL algorithms, i.e., policy-based, value-based, policy and value-based, respectively.

\subsection{Multi-agent Setting }  \label{sec:5.1}

The action space has two actions $\{f, \theta\}$ which we assume to be independent given the current state $S$; hence we choose a multi-agent (two-agent) cooperative setting in which both of these agents act to maximize the common reward. On the other hand, one can define a single agent that jointly predicts $\{f, \theta\}$. Results show that convergence and learning are much faster in the case of a multi-agent compared to a single-agent in both Vanilla Policy Gradient and Advantage Actor-Critic. 

\begin{figure}[t]
    \centering
    \begin{minipage}{.6\linewidth}
    \centering
    \includegraphics[width=\linewidth]{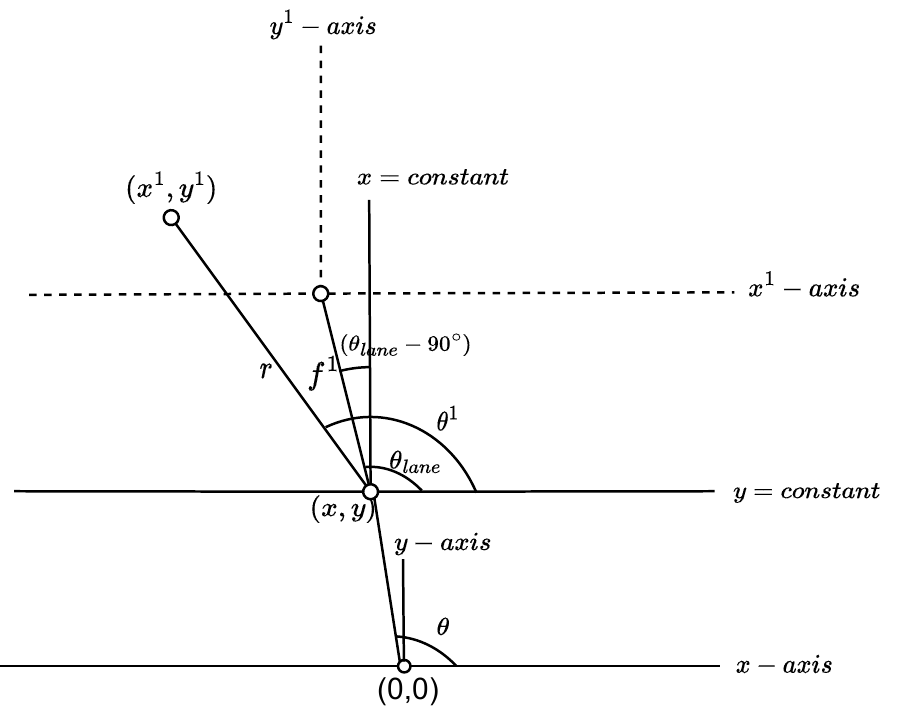}
    \captionof{figure}{Executing action. Here $x-axis, y-axis$ represents old coordinate axes, and $x^\prime-axis, y^\prime-axis$ represents new coordinate axes}
    \label{fig:Figure 4}
    \end{minipage}%
    \begin{minipage}{.35\linewidth}
    \centering
      \includegraphics[width=0.8\linewidth]{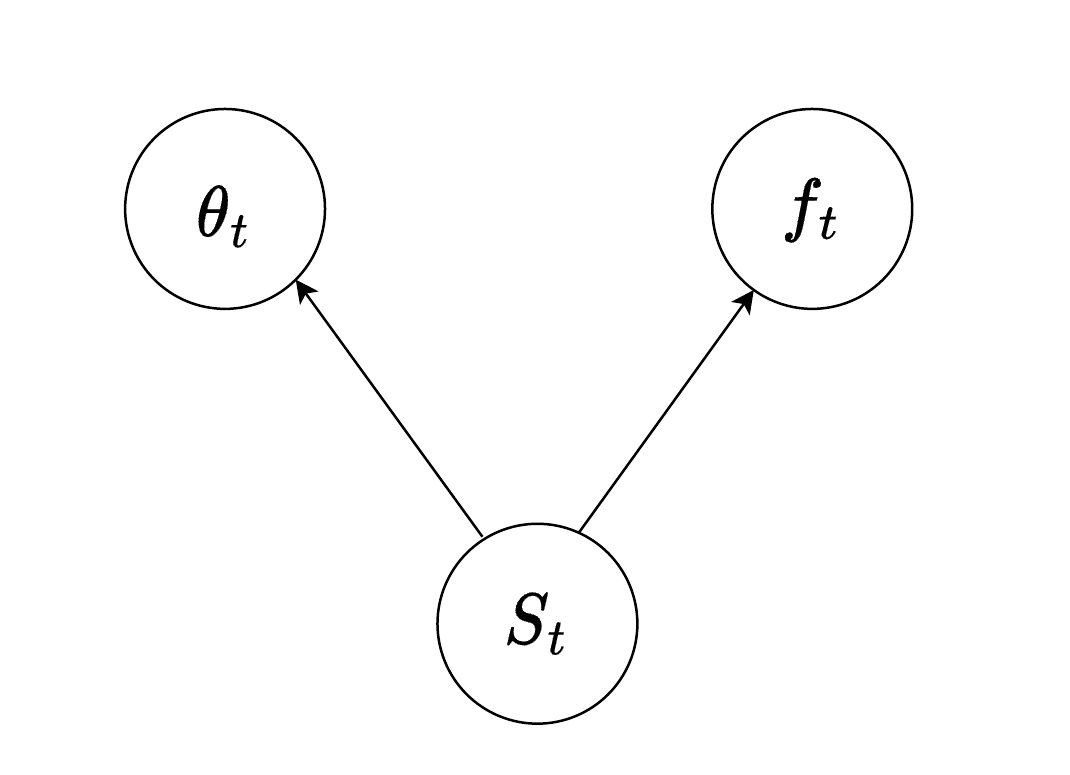}
    \captionof{figure}{Naive Bayes Model representation of $\theta_t$ and $f_t$ given state $S_t$}
    \label{fig:Figure_NB }
    \end{minipage}%
\end{figure}

 Consider the trajectory $\tau$ $\sim$ $\pi_{\phi_1,\phi_2}$ i.e.,  $\{S_0, a_0, S_1 ,a_1, S_2, a_2 \dots \}$ where action $a_t = (f_t , \theta_t)$. Actions at $t^{th}$ timestep $t$,  $f_t$ and $\theta_t$\ have a  correlation(statistically dependent), but from a causal view (Figure:\ref{fig:Figure_NB }) it is the third factor the state $S_t$(confounder or cause) that affects both actions independently. From lowering the speed and altering the angle on a curved lane to maintaining  the speed and angle on straight lanes concludes to have a correlation between speed and angle since they appear to move together(seeing $\theta_t$\ allows to predict $f_t$, but $f_t$ is not caused by $\theta_t$), however, it is the variables in the state space $S_t$ such as lane angle $\theta_{lane_{t}}$, etc, which causes both of these actions. However, actions $\theta_t$ and $f_t$ depend on both of the previous actions $\theta_{t-1}$ and $f_{t-1}$ since these variables constitute to state $S_t$. To preserve this causal relationship at timestep $t$, we claim that given state $S_t$ actions $f_t$ and $\theta_t$\ are independent. So, define two actor networks $\pi_{\phi_1}$, $\pi_{\phi_2}$ where $\phi_1$ and $\phi_2$ are parameters of speed network and angle network respectively.

Define $\prob(\tau;\phi_1;\phi_2)$ to be the probability of finding trajectory $\tau$ with policy $\pi_{\phi_1,\phi_2}$. We write,
    \begin{dmath} 
	 \prob(\tau;\phi_1;\phi_2)  = \prob(S_0)  \prod_{t=0}^{\infty} \pi_{\phi_1,\phi_2}(f_t,\theta_t|S_t)  \prob(S_{t+1}|S_t,f_t,\theta_t) \nonumber
	 \label{eq:eq3}
	 \end{dmath}
	 
	 Since given $S_t$, $f_t$ and $\theta_t$ are independent, we get
     \begin{dmath}
     \prob(\tau;\phi_1;\phi_2)= \prob(S_0) \prod_{t=0}^{\infty} \pi_{\phi_1}(f_t|S_t) \pi_{\phi_2}(\theta_t|S_t) \prob(S_{t+1}|S_t,f_t,\theta_t) \nonumber
     \label{eq:eq4}
     \end{dmath}

Define $J(\phi_{1};\phi_{2})$ to be the expected discounted reward with policy $\pi_{\phi_1,\phi_2}$ and $G(\tau)$ be the discounted cumulative reward obtained by trajectory $\tau$. We write, 
   \begin{dmath} 
   J(\phi_{1};\phi_{2})  =  \mathbb{E}_{\tau \sim \pi_{\phi_1,\phi_2}} \left[ \sum_{t=0}^{\infty} \gamma^{t} r_{t+1}| \pi_{\phi_1,\phi_2} \right]\\
    =  \sum_{\tau \sim \pi_{\phi_1,\phi_2}}\prob(\tau;\phi_1;\phi_2)G(\tau)\\
    \label{eq:eq5}
   \end{dmath}

\textbf{A. Vanilla Policy Gradient (Vanilla PG)} is an on-policy, policy-based, offline, model-free algorithm. To reduce the variance we took $G(\tau) = G_{t:\infty}(\tau) - b(S_t)$ where $G_{t:\infty}(\tau)$  is discounted cumulative reward from timestep $t$ and  $b(S_t)$ is the baseline which we took to be constant 
. Here, we propose \textbf{Dual-Actor Policy Gradient} architecture based on the eq:\ref{eq:eq4}, eq:\ref{eq:eq5}. We independently update both actors  $\phi_1$ and $\phi_2$ to maximize $J$  using gradient ascent where gradients are calculated as (proof in Supplementary Material)

	
    
       \begin{dmath} 
    \nabla_{\phi_1} \log \prob(\tau;\phi_1;\phi_2) 
    = \sum_{t=0}^{\infty} \nabla_{\phi_1} \log (\pi_{\phi_1}(f_t|S_t)) \nonumber
    \label{eq:eq6}
	\end{dmath}
	
	\begin{dmath}
	   \nabla_{\phi_{1}}J(\phi_{1};\phi_{2})  
     = \mathbb{E}_{\tau \sim \pi_{\phi_1,\phi_2}} \left[\sum_{t=0}^{\infty} \nabla_{\phi_{1}} \log (\pi_{\phi_{1}}(f_t|S_t)) G(\tau) \right] \\
     \label{eq:eq7}
    \end{dmath}

	Similarly, we get 


    	\begin{dmath}
       \nabla_{\phi_{2}} J(\phi_{1};\phi_{2})   =  \mathbb{E}_{\tau \sim \pi_{\phi_1,\phi_2}} \left[  \sum_{t=0}^{\infty} \nabla_{\phi_{2}} \log (\pi_{\phi_{2}}(\theta_t|S_t)) G(\tau) \right]
       \label{eq:eq8}
	\end{dmath}
	
\textbf{B. Advantage Actor Critic (A2C)} is an on-policy, policy, and value-based algorithm. We choose the baseline $b(S_t)$ to be state dependent expected return $ \mathbb{E}_{\tau \sim \pi_{\phi_1,\phi_2}} [ r_{t+1} + \gamma \cdot r_{t+2}+\dots| S_{t} = s  ] =  V^{\pi_{\phi_1,\phi_2}}(s)$.  We approximate the advantage function $A^{\pi_{\phi_1,\phi_2}}$($S_t,f_t ,\theta_t) = Q^{\pi_{\phi_1,\phi_2}}(S_t,f_t ,\theta_t) - V^{\pi_{\phi_1,\phi_2}}(S_t)$ with $r_{t+1}+\gamma \cdot V_{\psi}(S_{t+1})-V_{\psi}(S_{t})$ where $\psi$ is the parameters of the Value network $V_{\psi}$ which is used to approximate $V^{\pi_{\phi_1,\phi_2}}$. Here, we propose \textbf{Dual-Actor Single-Critic} architecture based on the \ref{eq:eq4}, \ref{eq:eq5}. We independently update both actors $\phi_1$ and $\phi_2$ to maximize $J$ in the same fashion as vanilla policy gradient. We update critic $\psi$ by minimizing the mean square error with the Bellman update equation. The gradients are calculated as (proof in Appendix section) \\

	    \begin{dmath} 
	   \nabla_{\phi_{1}}J(\phi_{1};\phi_{2})  
	      = \mathbb{E}_{\tau \sim \pi_{\phi_1,\phi_2}} \left[\sum_{t=0}^{\infty}   \nabla_{\phi_{1}} \log (\pi_{\phi_{1}}(f_t|S_t)) A^{\pi_{\phi_1,\phi_2}}(S_t,f_t,\theta_t)\right] \\
	      \label{eq:eq9}
	\end{dmath}

     Similarly, we get
    \begin{dmath} 
      \nabla_{\phi_{2}} J(\phi_{1};\phi_{2})  =  \mathbb{E}_{\tau \sim \pi_{\phi_1,\phi_2}} \left[  \sum_{t=0}^{\infty} \nabla_{\phi_{2}} \log (\pi_{\phi_{2}}(\theta_t|S_t)) A^{\pi_{\phi_1,\phi_2}}(S_t,f_t,\theta_t) \right]
      \label{eq:eq10}
    \end{dmath}
    
\textbf{C. Dueling Deep Q Networks (Dueling DQN)} is an off-policy, value-based method which is an extension over the standard DQN \cite{mnih2013playing}. Similar to A2C, we choose a single critic i.e., action value (critic) network $Q_{\psi}$ to approximate $Q^{\pi_{\phi_1,\phi_2}}$($S_t,f_t,\theta_t$). For a network $\psi$, Q value is calculated as $Q^{\pi_{\phi_1,\phi_2}}$($S_t,f_t ,\theta_t$) $\sim$ $V_{\psi}(S_t)+A_{\psi}(S_t,f_t,\theta_t)- \frac{1}{|A_{\psi}|}\sum_{f_t,\theta_t}(A_{\psi}(S_t,f_t,\theta_t))$. We update $\psi$ minimizing the mean square error in a similar fashion to DQN.

\subsection{Training}  \label{sec:5.2}

\textbf{A. Initialization: }  
We vary the initialization of the agent's center position $x$ and the starting angle $\theta$ following a uniform distribution and ensure that the agent is initialized within the lane limits $l_{lane}$ and $r_{lane}$. We assume that the agent always starts from rest, i.e., $f = 0$. These initializations are applicable every time we start an episode, ensuring the agent gets better at generalizing the environment.

\textbf{B. Data Augmentation: }
The major challenge for RL in an autonomous driving scenario is sampling efficiency. The learning process as a whole requires a huge number of samples to learn a reasonable policy because of an unbalanced distribution of observations from a larger state space. To overcome this, we divide a single video source into multiple short videos using a sliding window technique with fixed window size and stride ($0 < $stride $\leq $ window size), which we refer to as chunks of that video. This sort of data augmentation technique allows the agent to play numerous short games rather than one long game which makes the agent more robust as it learns or generalizes from multiple short games with various initializations. Also, when an agent plays a series of short games instead of one long game, it may perceive all the challenges present in the environment and can go beyond a particular transition.

\textbf{C. Train and Test:} We label a subset of videos from the chunks obtained from a single video as train chunks and the rest as test chunks. We train the agent over the train chunks. This ensures we haven't revealed the entire environment information to the agent. When testing, we test it on the entire single video (train and test chunks combined) with a fixed initialization. Test reward  Table:\ref{tab:Table 2} shows that the agent learns to generalize even on unseen chunks.




\section{Control: Execute the Action } \label{sec:6}

\subsection{ Motion }  \label{sec:6.1}

Actions $f^\prime$ and $\theta^\prime$ predicted in state $S_t$ transition the agent to state $S_{t+1}$.
 Predicted speed $f^\prime$ shifts the frames while predicted angle $\theta^\prime$ shifts the agent's position. Let $I$, $I^\prime$ be the frames in $t^{th}$, $(t+1)^{th}$ timesteps respectively. Considering how we've defined speed, we get $I^\prime = I$ + $f^\prime$

Let  $(x,y)$ be the agent's center in the frame $I$. Let ($x^\prime$,$y^\prime$) be the agent's new center after executing the action with reference to frame $I$. Let the distance between $(x,y)$ and ($x^\prime$,$y^\prime$) be $r$. However, due to the predicted speed $f^\prime$, frame $I$ shifts to $I^\prime$. So, we want to get the agent's new center with respect to the new frame $I^\prime$ and denote it by $(x^{\prime\prime},y^{\prime\prime})$. However, note that video frames(recorded by humans) always move with an angle  $\theta_{lane}$. We write, Figure:\ref{fig:Figure 4}

\begin{equation} 
\begin{split}
    x^\prime & = x + r \cdot \cos(\theta^\prime) \\
    y^\prime &= y + r \cdot  \sin(\theta^\prime)  
    \label{eq:eq11}
\end{split}
\end{equation}
\begin{equation} 
\begin{split}
    x^{\prime\prime} &= x^\prime - f^\prime \cdot  \sin(\theta_{lane}-90^\circ)  \\
    y^{\prime\prime} & = y^\prime - f^\prime \cdot \cos(\theta_{lane}-90^\circ) 
    \label{eq:eq12}
\end{split}
\end{equation}

Solving eq:\ref{eq:eq11} and eq:\ref{eq:eq12} gives  $x^{\prime\prime} = x + r \cdot  \cos(\theta^\prime) + f^\prime \cdot  \cos(\theta_{lane})$ and $y^{\prime\prime} = y + r \cdot  \sin(\theta^\prime) - f^\prime \cdot \sin(\theta_{lane})$. 

    \begin{table*}[t!]  
     \centering
      \begin{footnotesize}  
     \begin{tabular}{ |c|c|c|c|c|c|c|c| }
     \hline
     Data  & PG-MA & PG-SA & A2C-MA & A2C-SA & Dueling DQN-SA& Random-agent & Intelligent-agent \\
     \hline
     V1  & $14.11\pm9.36$  & $20.09\pm15.40$ & $8.75\pm12.57$ & $6.32\pm$8.93 & $22.04\pm21.08$ & $-1.06\pm4.09$ & $8.94\pm14.51$ \\
      V2  & $62.52\pm22.04$  & $44.07\pm14.35$ & $20.57\pm14.45$ &  $24.05\pm12.38$ &$40.63\pm32.49$ & $-0.33\pm3.54$ & $18.84\pm11.66$ \\
      V3  & $47.93\pm19.12$  & $8.09\pm2.05$ & $16.76\pm12.65$ & $13.05\pm6.83$ & $22.65\pm15.53$ & $-0.85\pm3.43$ & $15.99\pm10.87$ \\
     \hline
     \end{tabular}
     \caption{Mean $(\pm$standard deviation) of the reward for the last 50 episodes during training for videos V1, V2, and V3 for the proposed algorithms. Taking mean and standard deviation into consideration, PG-MA, and Dueling DQN seems to be better in all the 3 video environments. Here MA, and SA indicates Multi-agent and Single-agent setting respectively}
     \label{tab:Table 1}
     \end{footnotesize}
 \end{table*}
  
    \begin{table*}[t!]  
     \centering
      \begin{footnotesize}  
     \begin{tabular}{ |c|c|c|c|c|c|c|c|c| }
     \hline
     Data &Frames & PG-MA & PG-SA & A2C-MA & A2C-SA & Dueling DQN-SA & Random-agent & Intelligent-agent \\
     \hline
       V1  &983 & 182.91  & 300.54 & 109.49 & 76.51 &180.9 & -26.88 &137.05  \\
      V2  &1800 & 1380.37  & 1176.96 & 542 & 524.14&1257.68  &-32.59  & 480.21 \\
     V3  &1800 & 1184.71 & 212.16&437.33 & 260.73 & 549.72 & -27.05 & 408.67 \\
     \hline
     \end{tabular}
     \end{footnotesize} 
     \caption{Total test reward on all the chunks of the videos V1, V2, and V3 for the proposed algorithms. The total test reward depends on the number of frames within a video, the initialization used. We used $f = 0, \theta = 90^\circ, x = 0.5 \cdot (l_{lane} + r_{lane})$}.
     \label{tab:Table 2}
 \end{table*}

With the agent as a frame of reference(with the observer moving with an agent), the agent is at rest in $y-axis$ (y-coordinate is constant), it is the video frames that are coming toward the agent and it reacts by taking action. The video frame rate is controlled by the speed of the agent. Throughout this paper, the agent coordinate system is the frame of reference. Since we shift the coordinate axes with frame rate $f^\prime$ and hence $y$ coordinate before to be equal to $y$ coordinate after, we solve for $r$ in $y^{\prime\prime} = y$ giving $r = \frac{f^\prime \cdot    \sin(\theta_{lane})}{\sin(\theta^\prime)}$. Replacing the value of $r (r \geq 0)$, we get $x^{\prime\prime} = x + \frac{f^\prime \cdot  \sin(\theta^\prime + \theta_{lane})}{\sin(\theta^\prime)}$ .
Hence new coordinates with respect to new frame $I^\prime$ are $(x^{\prime\prime} , y^{\prime\prime}) = (\lfloor x +  \frac{f^\prime\cdot \sin(\theta^\prime+\theta_{lane})}{\sin(\theta^\prime)}\rfloor, y )$ 

Finally, state $S_{t+1} = $[features from concatenated depth map and segmentation map obtained from frame $I^\prime$, $\hat{c}_1 \cdot x^{\prime\prime}$, $ \hat{c}_2\cdot f^\prime$ , $\hat{c}_3\cdot \theta^\prime$,  $\hat{c}_4\cdot {l_{lane_{t+1}}}$, $\hat{c}_5\cdot {r_{lane_{t+1}}}$, $\hat{c}_6\cdot {\theta_{lane_{t+1}}}$  ]

\subsection{Frame of Reference: Simulator Vs Real-world Control }
In a simulator environment, we get the state as input from the simulator, learned policy predicts an action, the simulator executes the action and returns us the next state, the reward for that transition. In a simulator, we get the next frame based on the agents' predicted speed and angle. 

In real-world videos, we only have control over the frame rate $f$ since we can only move forward or backward within a video. The speed $f$ predicted by the agent makes the shift in video frames; however, the angle $\theta$ predicted by the agent causes the shift in the agent's position $(x,y)$ but not in the frames since we get the next frame based on the agents' speed, speed and angle of the human who drove the vehicle, resulting in the video data. To summarize, the simulator's motion is absolute, whereas, in real-world videos, the motion is relative to a human who drove the vehicle. 

This training framework makes the agent's motion relative to the training videos. But, in real-world deployment, the agent doesn't have reference videos to move. Hence we modify the motion during the inference to adapt to real-world use cases. Nevertheless, actions predicted by the algorithm are absolute and can work during real-world deployment. The only thing that differs in the deployment phase is unlike in training, the next state is based only on agents' actions. During the inference phase in deployment, after executing action through accelerator and steering, in the next state $S_{t+1}$, $I^\prime$ becomes the input camera frame (here this frame is dependent on both predicted $f, \theta$). Other features in state space remain unaltered.


\section{Experimental Results and Discussion} \label{sec:7}
 
    \subsection{Data \& Experimental Setup: }We have randomly chosen 3 driving videos from Youtube Figure:\ref{fig:Figure 1}. Videos V1 and V2 were recorded during the day, while video V3 was at night. Video V1 is a straight lane, while videos V2 and V3 are curved lanes. We took frame size $w = 224$ and resize every frame in all the videos to $(224 \times 224)$. Experiments show that the agent performs well irrespective of straight or curved lanes, day and night; this implies no bias factor introduced due to environmental changes or lane changes. We divide each of these videos into chunks and train the RL agent on the train chunks with the proposed algorithms, and initialization for all these 3 videos individually.

      \subsection{Baselines: }Since this work introduces discrete control in real-world environments, and has not been addressed in the literature there are no standard baselines available to compare against. We use the following to evaluate our agent:   
    \begin{enumerate}



 
    \item Compare the RL agent with random-agent, which chooses the actions from a uniform distribution.
    \item Compare the RL agent with an intelligent-agent that works on a simple action choice using if else condition to output angle $\theta$ given $x, l_{lane}, r_{lane}, \theta_{lane}$. We always take speed $f$ to be $\frac{f_{max}}{2}$.

    \item Observe the rolling average of the total mean reward over the last $N$ episodes. In our experiments, we took $N$ to be 50.
    \end{enumerate}
    
\subsection{Results and Discussion: }
\begin{figure*}[t]
    \begin{tabular}{cccc}
    \subfloat[Video V1]{{\includegraphics[width=8.5cm]{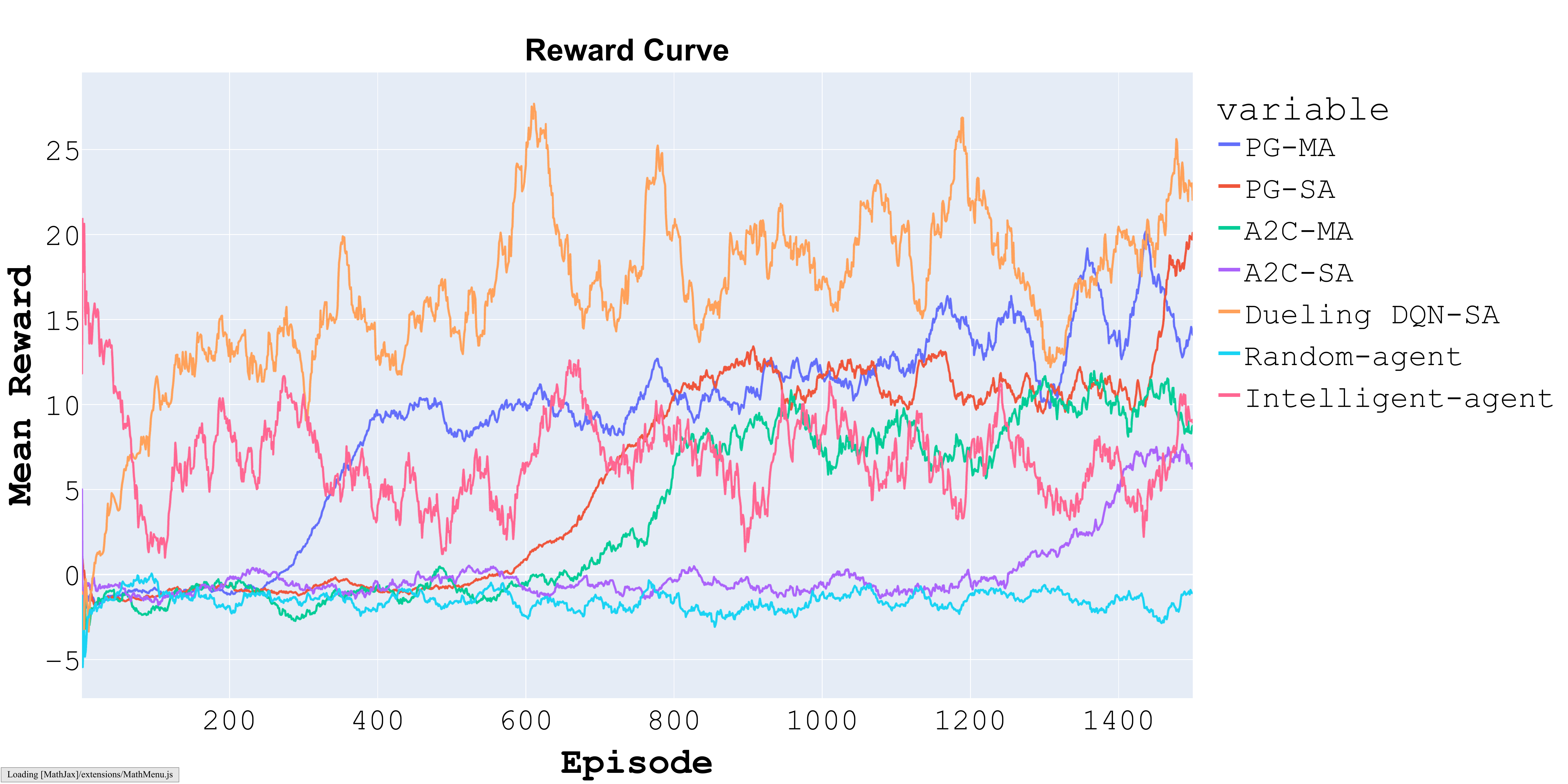} }} &
    \subfloat[Video V2]{{\includegraphics[width=8.5cm]{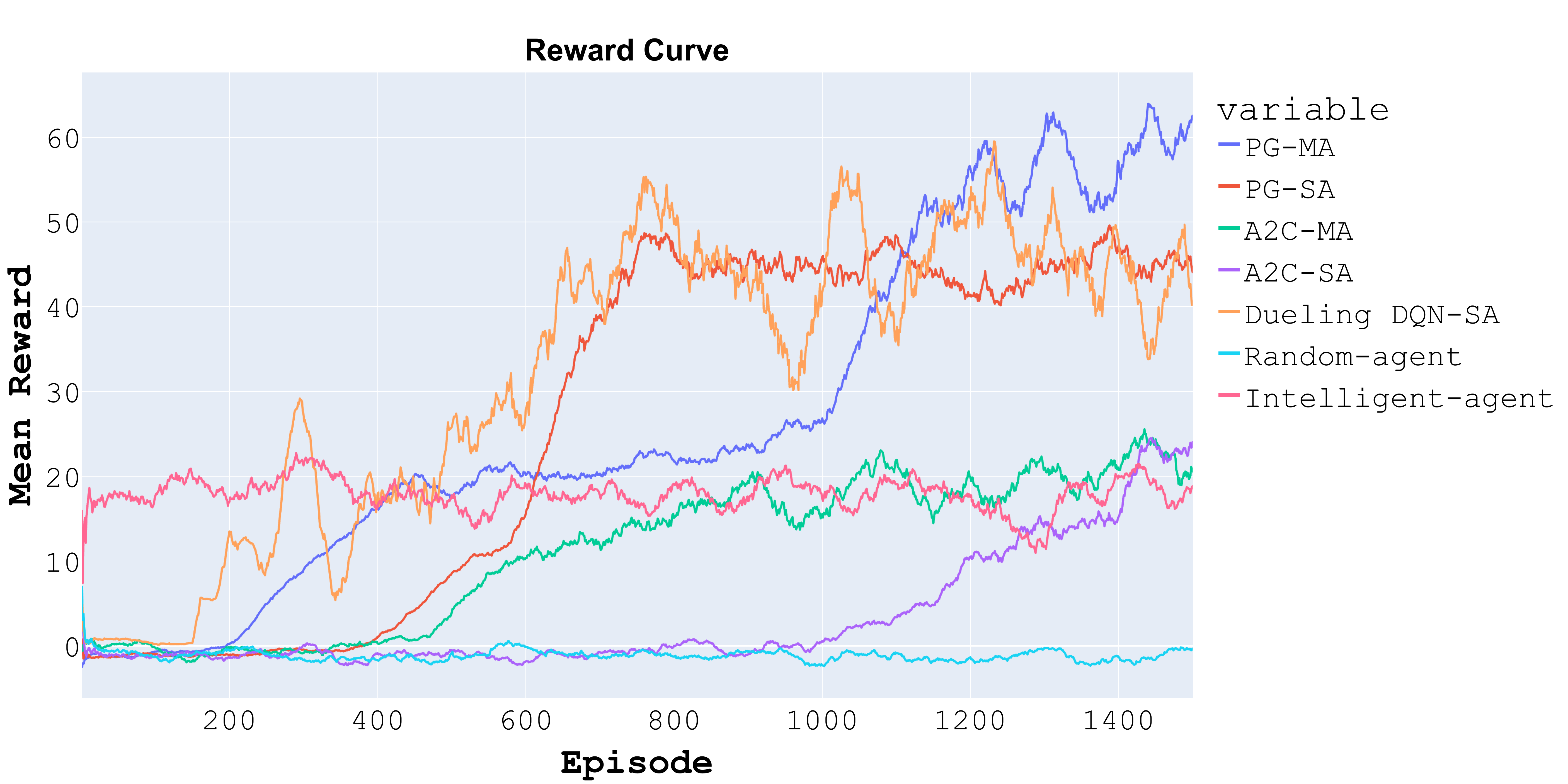} }} \\
    \subfloat[Video V3]{{\includegraphics[width=8.5cm]{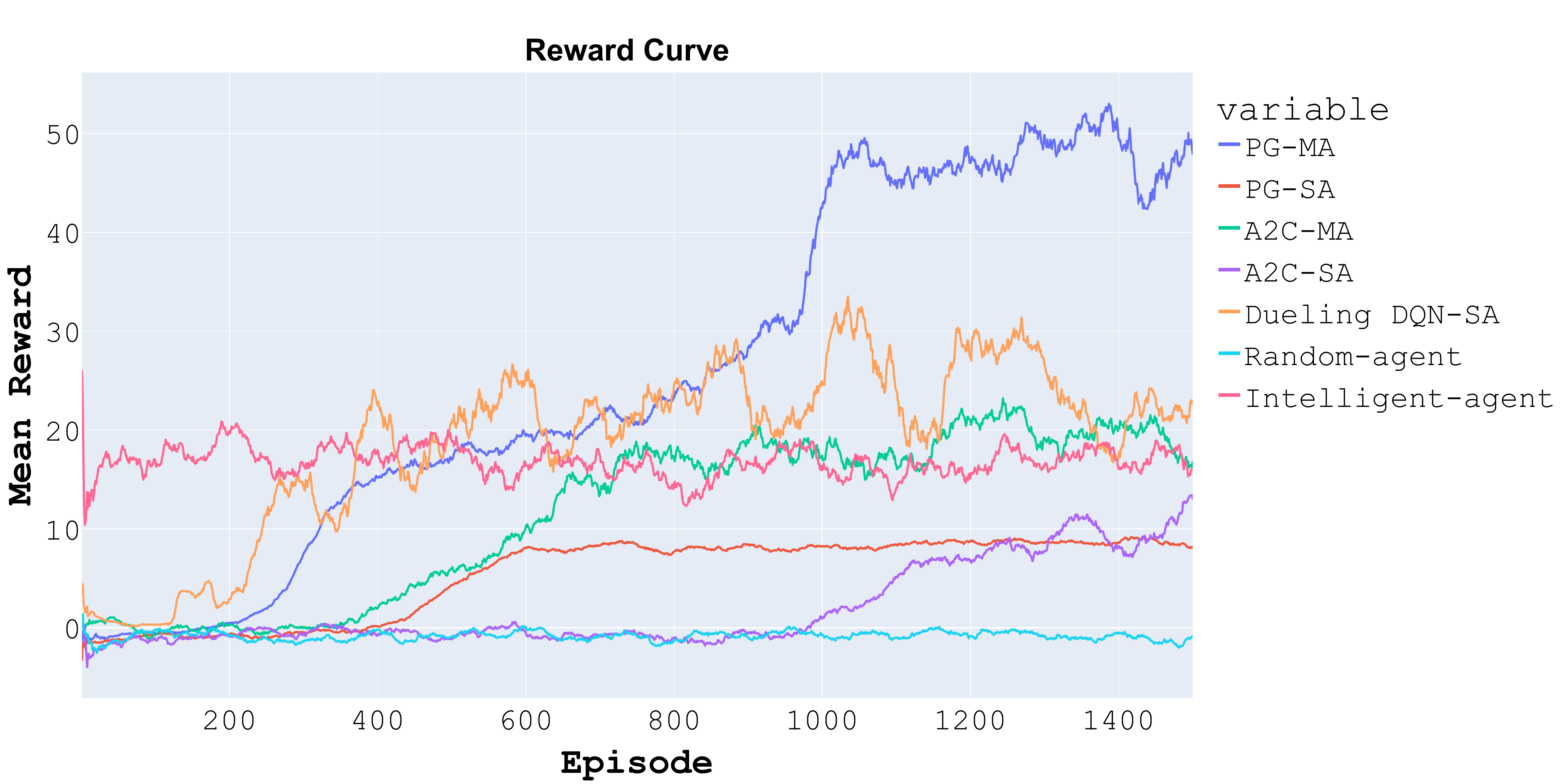} }} & 
    \subfloat[TORCS]{{\includegraphics[width=8.5cm]{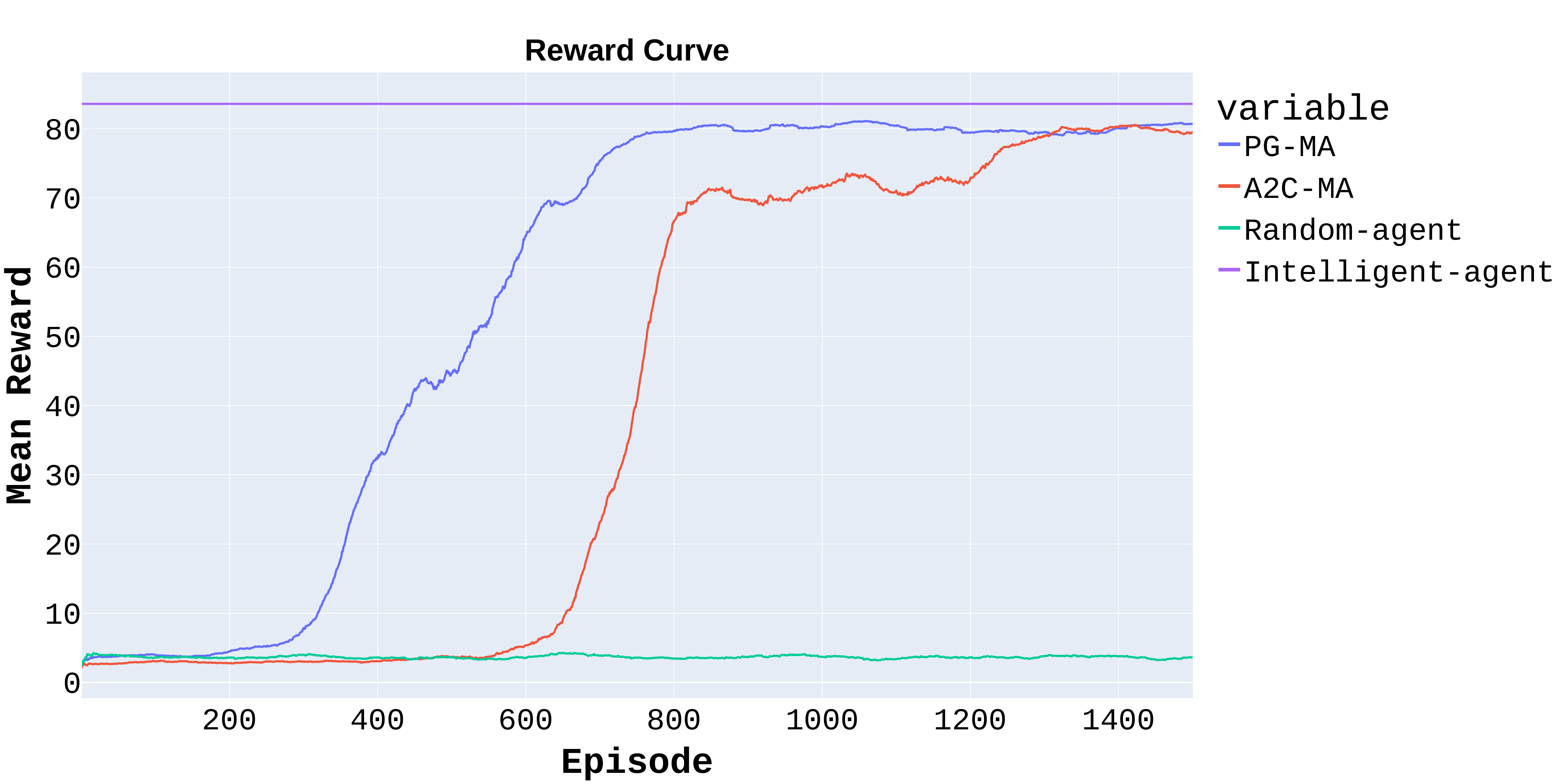} }}
    \end{tabular}

    \caption{Training rewards on environments V1, V2, V3, and TORCS. Multi-agent setting in Vanilla PG, A2C made the learning faster and achieved better performance compared to single-agent setting.}
    \label{fig:Figure 5}
    \end{figure*}

We ran Vanilla PG, A2C for both single-agent and multi-agent settings, Dueling DQN for single-agent settings, random-agent, and intelligent-agent for all the 3 video environments. Results Figure:\ref{fig:Figure 5} demonstrate that the reward function increases significantly even with fewer episodes of training. This implies that the RL agent is capable of learning even with minimal training(1500 episodes is relatively small), and minimal data.

 Table:\ref{tab:Table 1}, Table:\ref{tab:Table 2} show that Vanilla PG, Dueling DQN, and A2C outperforms random and intelligent agent in all 3 environments. However, Vanilla PG and Dueling DQN outperform A2C in all the 3 environments because batch update and replay memory played a crucial role in the learning process of Vanilla PG and Dueling DQN respectively. We ran Vanilla PG with batch sizes $(1,3,5)$. Batch size 5 showed significant improvement in reward. We use replay memory of 5000 with batch size 128 in Dueling DQN. 
 
 We explain the better performance and faster learning of Vanilla PG and A2C in multi-agent settings compared to single-agent because of the causal view that we took into consideration. An AI model which captures causal relationships of data is one way to ensure that the model will generalize to a new setting. Moving from the model that captures correlation to the one that captures causal behavior is hence recommended.  
 
    For all of the experiments, we took discount factor  $\gamma = 0.99$, the learning rate for both agents to be $5\times 10^{-6}$. In reward formulation, we took $\alpha$, $\beta$, $\mu$, $\nu$ to be 1, and $\delta$ to be 0.1,  the number of episodes to be at-max 1500. We took the number of frames in a chunk or window size to be 100. We have used the PyTorch framework, Ubuntu version 18.04 to run the experiments.

      \subsection{TORCS:} We argue that the real-world environment is a superset over the virtual environment; hence, the methods which apply to the virtual environment might not apply to the entire real space since it requires a lot more understanding, but the vice-versa is true. With only slight modifications, we can transfer real-world knowledge to virtual environments. Hence to test the efficacy of our method in a virtual environment, we have deployed it in TORCS. TORCS is an open-source simulator specially used for racing. 
      
      
      Discrete TORCS environment \cite{Torcs} (Michigan with uni-directional road) has a action space $\mathcal{A} = [accel, brake, steer]$ where $accel \in \{0,1\}, brake \in \{0,1\}, steer \in \{-1,0,1\} $ however $accel$ and  $brake$ $\neq$ 1 simultaneously, so the $|\mathcal{A}| = 9$.  The state space is the same as in the real-world, but here we took advantage of sensor values for speed, angle, and track info from the simulator. Hence we skipped the pre-processing steps that were defined for real-world and adjusted the reward accordingly and applied Vanilla Policy Gradient, Advantage Actor-Critic in a multi-agent setting on top of it to obtain the actions. Execution of the action is taken care of by the simulator itself. Even in the TORCS simulator, Vanilla PG multi-agent, A2C multi-agent outperforms random-agent and achieves an intelligent-agent level performance Figure:\ref{fig:Figure 5}. The hyper-parameters were the same as in the case of the real world, if applicable.

\section{Conclusion and Future Work} \label{sec:8}



This paper introduces discrete control in real-world environments by introducing perception, planning, and control in real-world driving environments. The RL agent learns with minimal video data, and minimal training using the proposed multi-agent setting and training procedures. The proposed methods will work in simulated environments as well. Future works may follow in the footsteps of this work. We incorporate GPS, LIDAR,  and other sensors into the state space and then train an agent with the help of the sensors and vision data. One can extend this work on bidirectional roads as well by reshaping the reward function appropriately to restrict the agent to be on one-half of the lane. This work can also be extended into continuous control by discretizing the continuous action space into several discrete values up to the necessary precision.





{\small
\bibliographystyle{unsrt}
\bibliography{Ref}
}


\newpage
\section{Appendix}

\subsection{Vanilla Policy Gradient in Multi-agent setting: } 

     \text{Consider, } 
     \begin{dmath} 
    \nabla_{\phi_1} \log \prob(\tau;\phi_1;\phi_2) = \nabla_{\phi_1} \log \left(\prob(S_0)\prod_{t=0}^{\infty} \pi_{\phi_1}(f_t|S_t) \pi_{\phi_2}(\theta_t|S_t) \prob(S_{t+1}|S_t,f_t,\theta_t)  \right)   
    \end{dmath}

       \text{Since $\prob(S_0)$ is independent of $\phi_1$, we write }
    \begin{dmath}
   \nabla_{\phi_1} \log \prob(\tau;\phi_1;\phi_2)  = \nabla_{\phi_1} \left[ \sum_{t=0}^{\infty} \log(\pi_{\phi_1}(f_t|S_t)) + \sum_{t=0}^{\infty} \log( \pi_{\phi_2}(\theta_t|S_t) )  +  \sum_{t=0}^{\infty} \log( \prob(S_{t+1}|S_t,f_t,\theta_t)) \right]\\
     = \sum_{t=0}^{\infty} \nabla_{\phi_1} \log (\pi_{\phi_1}(f_t|S_t)). \nonumber
    \end{dmath}

\text{Consider, }
    \begin{dmath}	
	   \nabla_{\phi_{1}}J(\phi_{1};\phi_{2})    = \nabla_{\phi_1} \left[\sum_{\tau \sim \pi_{\phi_1,\phi_2}}\prob(\tau;\phi_1;\phi_2)G(\tau)\right]  \\ 
	    = \sum_{\tau \sim \pi_{\phi_1,\phi_2}}\nabla_{\phi_1} \left[\prob(\tau;\phi_1;\phi_2)G(\tau)\right]  \\ 
	     = \sum_{\tau \sim \pi_{\phi_1,\phi_2}} \frac{\prob(\tau;\phi_1;\phi_2)}{\prob(\tau;\phi_1;\phi_2)} \nabla_{\phi_1} \left[\prob(\tau;\phi_1;\phi_2)\right] G(\tau) \\ 
	     = \sum_{\tau \sim \pi_{\phi_1,\phi_2}} \frac{\nabla_{\phi_1} \left[\prob(\tau;\phi_1;\phi_2)\right] }{\prob(\tau;\phi_1;\phi_2)} \prob(\tau;\phi_1;\phi_2) G(\tau) \\ 
	     = \sum_{\tau \sim \pi_{\phi_1,\phi_2}} \left[\nabla_{\phi_1} \log \prob(\tau;\phi_1;\phi_2) \right]\prob(\tau;\phi_1;\phi_2) G(\tau) \\ 
	   = \mathbb{E}_{\tau \sim \pi_{\phi_1,\phi_2}}  \left[\nabla_{\phi_1} \log \prob(\tau;\phi_1;\phi_2)  G(\tau) \right]  \\ 
     = \mathbb{E}_{\tau \sim \pi_{\phi_1,\phi_2}} \left[\sum_{t=0}^{\infty} \nabla_{\phi_{1}} \log (\pi_{\phi_{1}}(f_t|S_t)) G(\tau) \right] \\
    \end{dmath}

    \text{Similarly, we get }
	 \begin{dmath}
    \nabla_{\phi_2} \log \prob(\tau;\phi_1;\phi_2)  = \sum_{t=0}^{\infty} \nabla_{\phi_2} \log (\pi_{\phi_2}(\theta_t|S_t)) 
	\end{dmath} 

  \text{and }
    \begin{dmath} 
       \nabla_{\phi_{2}} J(\phi_{1};\phi_{2})   =  \mathbb{E}_{\tau \sim \pi_{\phi_1,\phi_2}} \left[  \sum_{t=0}^{\infty} \nabla_{\phi_{2}} \log (\pi_{\phi_{2}}(\theta_t|S_t)) G(\tau) \right]
	\end{dmath}
	
\subsection{Advantage Actor-Critic in Multi-agent setting: } 

        \begin{dmath}
        \text{Consider $\nabla_{\phi_{1}}J(\phi_{1};\phi_{2})$ and substitute $G(\tau) = G_{t:\infty}(\tau) - b(S_t)$, we get}\\
	   \nabla_{\phi_{1}}J(\phi_{1};\phi_{2})  = \mathbb{E}_{\tau \sim \pi_{\phi_1,\phi_2}} \left[\sum_{t=0}^{\infty} \nabla_{\phi_{1}} \log (\pi_{\phi_{1}}(f_t|S_t)) \left[ G_{t:\infty}(\tau) - b(S_t)\right] \right] \\
     \text{Using nested conditional expectation and substituting value of $b(S_t) = V^{\pi_{\phi_1,\phi_2}}(S_t)$, we get}\\
	      \nabla_{\phi_{1}}J(\phi_{1};\phi_{2})  \\ = \mathbb{E}_{\tau \sim \pi_{\phi_1,\phi_2}} \left[\sum_{t=0}^{\infty} \mathbb{E}_{\tau \sim \pi_{\phi_1,\phi_2}} \left[  \nabla_{\phi_{1}} \log (\pi_{\phi_{1}}(f_t|S_t))  G_{t:\infty}(\tau) | S_t,f_t,\theta_t \right] \right] -  \mathbb{E}_{\tau \sim \pi_{\phi_1,\phi_2}} \left[\sum_{t=0}^{\infty} \nabla_{\phi_{1}} \log (\pi_{\phi_{1}}(f_t|S_t)) V^{\pi_{\phi_1,\phi_2}}(S_t) \right]  \\
	      = \mathbb{E}_{\tau \sim \pi_{\phi_1,\phi_2}} \left[\sum_{t=0}^{\infty}   \nabla_{\phi_{1}} \log (\pi_{\phi_{1}}(f_t|S_t))  \mathbb{E}_{\tau \sim \pi_{\phi_1,\phi_2}}  \left[ G_{t:\infty}(\tau) | S_t,f_t,\theta_t \right]\right ]  -  \mathbb{E}_{\tau \sim \pi_{\phi_1,\phi_2}} \left[\sum_{t=0}^{\infty} \nabla_{\phi_{1}} \log (\pi_{\phi_{1}}(f_t|S_t)) V^{\pi_{\phi_1,\phi_2}}(S_t) \right]  \\
	      = \mathbb{E}_{\tau \sim \pi_{\phi_1,\phi_2}} \left[\sum_{t=0}^{\infty}   \nabla_{\phi_{1}} \log (\pi_{\phi_{1}}(f_t|S_t)) Q^{\pi_{\phi_1,\phi_2}}(S_t,f_t,\theta_t)\right]   -  \mathbb{E}_{\tau \sim \pi_{\phi_1,\phi_2}} \left[\sum_{t=0}^{\infty} \nabla_{\phi_{1}} \log (\pi_{\phi_{1}}(f_t|S_t)) V^{\pi_{\phi_1,\phi_2}}(S_t) \right]  \\
	      	\end{dmath}
	      	
Since $A^{\pi_{\phi_1,\phi_2}}(S_t,f_t,\theta_t) \triangleq Q^{\pi_{\phi_1,\phi_2}}(S_t,f_t,\theta_t) - V^{\pi_{\phi_1,\phi_2}}(S_t)$, we get 
\begin{dmath}
	    \nabla_{\phi_{1}}J(\phi_{1};\phi_{2})   
	      = \mathbb{E}_{\tau \sim \pi_{\phi_1,\phi_2}} \left[\sum_{t=0}^{\infty}   \nabla_{\phi_{1}} \log (\pi_{\phi_{1}}(f_t|S_t)) A^{\pi_{\phi_1,\phi_2}}(S_t,f_t,\theta_t)\right] \\
\end{dmath}
	
     Similarly, we get
    \begin{dmath} 
      \nabla_{\phi_{2}} J(\phi_{1};\phi_{2})  =  \mathbb{E}_{\tau \sim \pi_{\phi_1,\phi_2}} \left[  \sum_{t=0}^{\infty} \nabla_{\phi_{2}} \log (\pi_{\phi_{2}}(\theta_t|S_t)) A^{\pi_{\phi_1,\phi_2}}(S_t,f_t,\theta_t) \right]
    \end{dmath}

\end{document}